# Explainable Neuro-Fuzzy Prediction for Trustworthy Decision-Making in Maritime

Dionisis Kalogeropoulos
*Dept of Computer Science and Biomedical Informatics*
*University of Thessaly*
Lamia, Greece
dkalogerop@uth.gr

Georgia Sovatzidi
*Dept of Computer Science and Biomedical Informatics*
*University of Thessaly*
Lamia, Greece
gsovatzidi@uth.gr

Dimitris K. Iakovidis
*Dept of Computer Science and Biomedical Informatics*
*University of Thessaly*
Lamia, Greece
diakovidis@uth.gr

***Abstract*— Predicting when maritime systems require maintenance can be critical, avoiding hazards and costly consequences. To address this problem, this paper proposes an explainable decision-making framework that integrates a neuro-fuzzy prediction model with a two-stage explainable component. The first stage of this component produces feature-attribution explanations, using gradient-based saliency maps, and the second stage extracts local rules using a fuzzy decision tree. The proposed framework is generic and can be integrated into any deep learning-based approach, rendering it explainable. To the best of our knowledge, this is the first fuzzy logic-based framework enabling both feature-level and local rule-based explanations of black box models. This approach aims to foster trustworthiness in decision making through user-understandable machine inferences. The performance of the proposed framework using a deep residual-based neural backbone is evaluated on various general-purpose public benchmark datasets, and its utility in maritime is demonstrated in the context of early fault detection in a naval propulsion system dataset. The results indicate that it can provide predictions outperforming relevant state-of-the-art approaches, with an average AUC-ROC (Area Under the Receiver Operating Characteristic Curve) value, reaching up to 99%, while offering the advantage of explainability.**

## I. Introduction

Effective predictive maintenance strategies are critical to ensuring the operational efficiency, reliability, and longevity of marine vessels, especially those equipped with complex propulsion systems. In marine operations, predictive maintenance plays a critical role by continuously monitoring system health indicators and enabling maintenance actions to be performed early before failure, thereby reducing the probability of hazards, unnecessary downtime and consequent costs [1]. In this context, changes in the health status of a naval propulsion system (NPS) can be detected by identifying degradation states, which represent progressive levels of performance deterioration relative to normal operating conditions. These states capture the gradual transition from healthy to degraded operation and are inferred from sensor measurements [2]. Although degradation-state identification, early warning, and failure prediction may appear as distinct tasks, all can be formalized under a unified binary classification framework, which is the formulation adopted throughout this paper.

This work has received funding by the European Union (EU)'s Horizon Europe research and innovation programme under grant agreement No101202933 (D-NAVIO). Views and opinions expressed are however those of the authors only and do not necessarily reflect those of the EU or the European Climate, Infrastructure, and Environment Executive Agency (CINEA). Neither the European Union nor the granting authority can be held responsible for them.

Despite the importance of predictive maintenance for naval applications, only a few studies have focused on machine learning-based solutions facilitating early warning for deterioration of the system. In particular, Cipollini *et al.* [3] developed an approach that leveraged both supervised and unsupervised learning to model the performance degradation of key NPS components, including the gas turbine, compressor, hull, and propeller. Gao *et al.* [4] proposed the DRN-GAN framework that combined a Deep Residual Network with a Generative Adversarial Network to address data imbalance and nonlinear relations, enabling degradation assessment and performance prediction for naval gas turbines. Javadnejad *et al.* [5] introduced a hybrid probabilistic model, BiGMM-HMM that integrated a Bivariate Gaussian Mixture Model with a Hidden Markov Model to enhance predictive maintenance performance and capture degradation dynamics in naval propulsion equipment. Furthermore, Sahraoui *et al.* [6] applied ensemble learning to improve fault prediction accuracy in NPS maintenance. However, the existing approaches do not provide explainable predictions of their outcomes, thus limiting the trust of the users.

In this paper, we present a novel framework comprising a neural network-based (NN) fuzzy classifier that serves as the primary predictive model with a two-stage explainable component. To explain the predictions of the classifier, the explainable component first employs gradient-based saliency maps to identify input features that are most influential for a given decision. Secondly, a Fuzzy Decision Tree (FDT) trained over instance-specific local neighborhoods is used to extract fuzzy rules, providing explanations of the model's predictive behavior. The model is capable of automatically extracting rules directly from data, without requiring prior domain knowledge, thereby alleviating the reliance on manual rule engineering. The rest of the paper is organized into four sections. Section 2 details the proposed neuro-fuzzy prediction framework, section 3 presents the experiments and the results obtained, and section 4 provides examples focusing on the explainability of the framework in the context of failure prediction for NPSs. Conclusions derived from this study are summarized in section 5.

## II. Neuro-Fuzzy Prediction Framework

An overview of the proposed framework is presented in Fig. 1; a neural network classifier serves as the primary predictive model, while a fuzzy decision tree is learned as a surrogate to explain instance-specific areas with human-readable rules. Formally, let $\boldsymbol{x} \in \mathbb{R}^d$ denote an input vector.

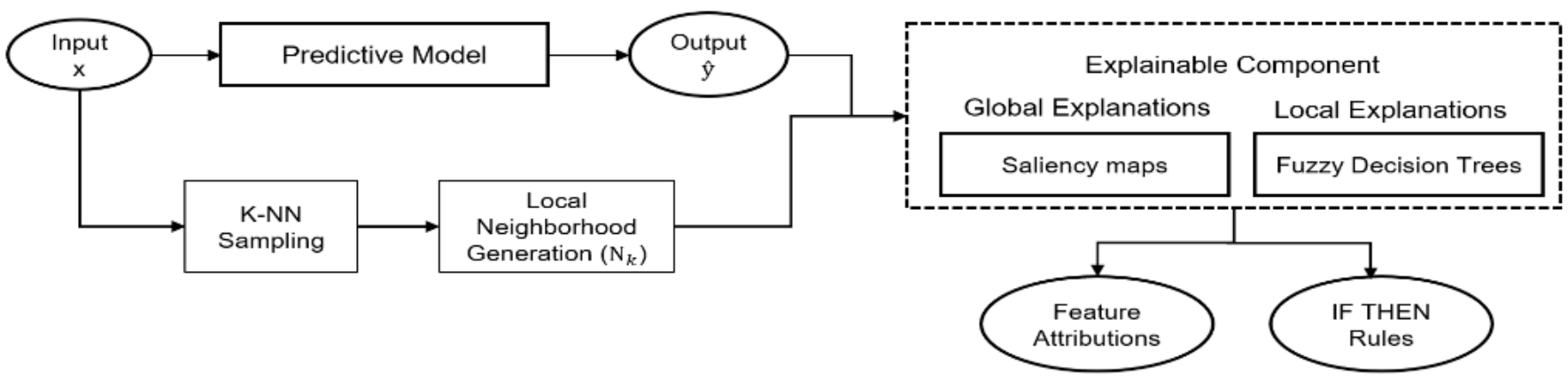

Fig 1. Overview of the proposed framework.

We consider a predictive model $T$ that defines a function $Y : \mathbb{R}^{d} \to \mathbb{R}^{C}$, where $C$ is the number of classes in the classification setting. In addition, we employ an explanation map $E : \mathbb{R}^{d} \to \mathbb{R}^{d}$, which assigns an importance score to each feature in the original feature space.

### A. Predictive Model

The architecture consists of a feature extractor followed by a fuzzy classifier head that serves as the primary predictive model. The feature extractor is constructed from sequential processing of blocks that incorporate linear transformations, batch normalization, dropout for regularization and residual connections that enhance training stability and optimization performance [8]. The fuzzy classifier head is designed to integrate fuzzy logic to the prediction process of the network by implementing a differentiable fuzzy membership function, following the paradigm of neuro-fuzzy systems, such as ANFIS [7] and its deep extensions [8], [9]. By modelling the NN's neuron overlap with degrees of membership rather than hard assignments, the classifier produces smoother decision boundaries that are robust to noise with better generalization. This is achieved by combining two layers, a fuzzification layer and a defuzzification layer. The first applies a learnable affine transformation composed of scaling, rotation, and translation parameters, followed by an $\ell_2$-norm and exponential mapping to obtain membership degrees. The defuzzification layer then combines these degrees to obtain the final model output. Specifically, the fuzzification layer implements a generalized membership function [9]:

$$\mu_{f_i}(\boldsymbol{u}) = \exp\left(-||\boldsymbol{A}_{f_i}\boldsymbol{u} + \boldsymbol{b}_{f_i}||_2\right) \tag{1}$$

where, $\boldsymbol{u} \in \mathbb{R}^h$ denotes the feature representation produced by the previous layer of dimensionality $h \in \mathbb{N}$, $f_i, i = 1, \dots F$, indexes the fuzzy rules and $F$ is the total number of rules, where $\mu_{f_i}(\boldsymbol{u})$ represents the membership degree of $\boldsymbol{u}$ to the i-th fuzzy rule. The matrix $\boldsymbol{A}_{f_i} \in \mathbb{R}^{h \times h}$, $f_i, i = 1, \dots F$, is a learnable linear transformation that controls the orientation and scaling of the fuzzy set in the latent space, $\boldsymbol{b}_{f_i} \in \mathbb{R}^h, f_i, i = 1, \dots F$, corresponds to the calculated centroids. The normalized transformed input is mapped to a value in $[0,1]$, using the exponential function. Moreover, the defuzzification process is implemented through a learnable linear layer that aggregates the activations of all fuzzy rules to produce a final crisp output. Specifically, the defuzzified output $\widehat{\boldsymbol{y}}_{\boldsymbol{T}} \in \mathbb{R}^c$ is computed as a normalized weighted sum of the learned rule vectors $\boldsymbol{z}_{f_i} \in \mathbb{R}^h$ [10]:

$$\widehat{\boldsymbol{y}}_{\boldsymbol{T}} = \frac{\sum_{i=1}^{F} \alpha_{f_i}(u)\, z_{f_i}}{\sum_{i=1}^{F} \alpha_{f_i}(u)} \tag{2}$$

where $\alpha_{f_i}(\boldsymbol{u})$ denotes the activation of fuzzy rule $f_i$.

### B. Explainable Component

Enabling trustworthy decision-making requires transparent methods capable of explaining their outcomes. We employ a two-stage explanation component that provides a ranking of the features that influenced the output. Also, it generates local fuzzy IF-THEN rules by training local FDTs.

**Feature-Attribution Explanations**

The proposed framework incorporates a gradient-based saliency mechanism to provide explanations through saliency maps. Saliency is computed by evaluating the gradient of the output of a model T with respect to the input features, thereby quantifying each feature's contribution to the prediction. The standard gradient explanation ($E_{grad}$) constitutes a feature-attribution explanation map that assigns an importance score to each feature in the original feature space, and is defined as follows [11]:

$$E_{grad}(x) = \left|\frac{\partial Y}{\partial x}\right| \tag{3}$$

The gradient quantifies the sensitivity of the predictions $Y(\mathrm{x})$ to small changes in each input dimension in an area around the input $\boldsymbol{x}$. Although computed with respect to the model output, the resulting explanations are inherently influenced by the structure of the network. By adding a fuzzy classifier, saliency maps are shaped by the fuzzy model structure itself. As a result, gradient-based saliency reflects both output sensitivity and the fuzzy reasoning, providing feature importance maps that explain the inferred predictions.

**Local Explanations**

To further enhance the explainable outcomes, we incorporate a second layer of explainability using a FDT that enables local, instance-based extraction of human-readable rules. FDTs allow soft (fuzzy) membership of samples to multiple branches, which improves robustness to noisy and uncertain measurements. Furthermore, employing soft splits leads to smoother decision boundaries and avoids abrupt changes in the decision function, while still producing interpretable human-readable rules, using linguistic terms, *e.g.,* Low, Medium, High [12]. In the proposed framework, the NN acts as the primary predictor: for each queried instance, we collect a local neighborhood around the query sample measured by a predefined distance metric along with the model's corresponding predictions, and fit a shallow FDT using Gaussian membership functions and a weighted fuzzy entropy split criterion [13], followed by pruning. Pruning is applied to remove low-support branches, reducing overfitting in the local neighborhood. The trained tree is then translated into a set of fuzzy "IF–THEN" rules that explains the predi-ction of $T$ in the original feature space, within a localized instance-specific setting, thereby providing an additional explanation dimension.

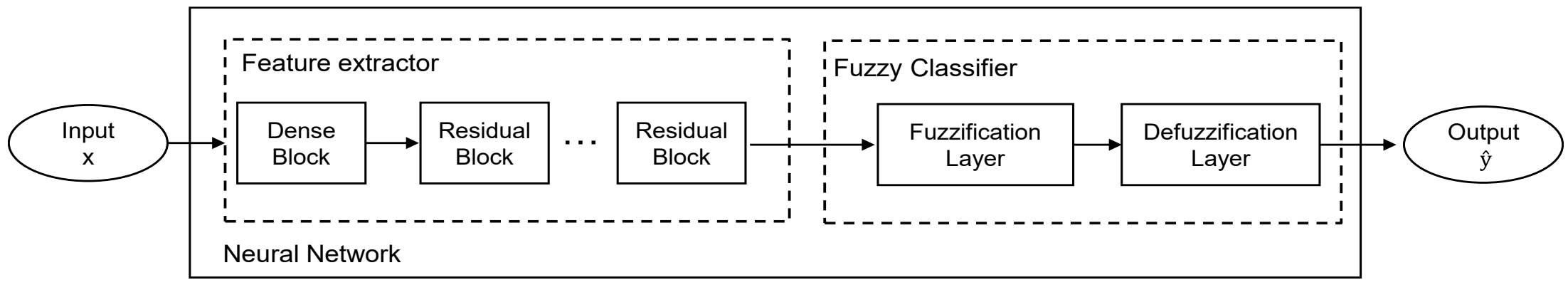

Fig 2. The proposed neural network architecture that integrates a fuzzy classifier, used as the predictive model (Fig 1).

### C. Inference Step

During the inference step (Fig. 1), an input sample $\boldsymbol{x}$ is processed by the model $T$ to produce an output $\widehat{\boldsymbol{y}}_{\boldsymbol{T}}$ (Fig. 2). A two-stage explainable component is then applied to generate explanations for this output. In the first stage, gradient-based saliency is computed by evaluating the gradient of $\widehat{\boldsymbol{y}}_{\boldsymbol{T}}$ with respect to the input features. The resulting saliency vector $\boldsymbol{E}_{\boldsymbol{grad}}$ captures the magnitude of the contribution of features influencing $\widehat{\boldsymbol{y}}_{\boldsymbol{T}}$ providing a feature-level explanation of the predictive model's decision. In the second stage, a stratified local neighborhood is constructed around the input sample $\boldsymbol{x}$, using k-NN. The sampled neighborhood $\boldsymbol{N}_{\boldsymbol{k}}$ and the model predictions $\widehat{\boldsymbol{y}}_{\boldsymbol{k}}$, form the local training dataset $\{\boldsymbol{N}_{\boldsymbol{k}}, \widehat{\boldsymbol{y}}_{\boldsymbol{k}}\}$ which is used to train the FDT. This neighborhood defines a local region of the feature space around $\boldsymbol{x}$, which within the FDT defines fuzzy sets that cover overlapping regions of each feature in $\boldsymbol{N}_{\boldsymbol{k}}$. The centers of these sets are initialized using linear interpolation between the local extrema of the corresponding feature values. Based on this locally trained FDT with $m$ total nodes, each unique path from the root to a leaf node constitutes a fuzzy rule leading to a class prediction $K$. In the proposed framework, final predictions are calculated by aggregating the firing strengths of the activated tree nodes. For nodes representing features $f_1, \ldots, f_m$ explanations come as IF-THEN local rules of the form:

**IF** ($f_1$) is *Low* **AND** ($f_2$) is *High* **THEN** ($\widehat{\boldsymbol{y}}$) belongs to $K$.

Each rule is associated with a support value $s$ (4) and a confidence value $p$ (5), which quantify the local precision of the rules. Support measures the extent to which a rule is activated across the dataset, computed as the firing strength of the rule over all samples. Confidence indicates the reliability of the linguistic antecedent's ability to map inputs to the target label.

$$s(R_i) = \sum_{x \in \mathcal{D}} \alpha_{c_i}(x) \quad (4)$$

$$p(R_i) = \frac{\sum_{x \in \mathcal{D}} \alpha_{c_i}(x|\hat{y} = K)}{\sum_{x \in \mathcal{D}} \alpha_{c_i}(x)}, \quad i = 1, \ldots, \Phi \quad (5)$$

Each rule $R_i$, corresponds to a root-to-leaf path in the FDT and is associated with a fuzzy region of the feature space, with $\Phi \in \mathbb{N}$ denoting the total number of rules. Let $a_{c_i} \in [0,1]$ denote the firing strength (activation) of rule $R_i$ for an input $\boldsymbol{x}$, defined as the aggregation of the membership degrees along the corresponding path, where $\boldsymbol{x} \in \mathcal{D}$ and $\mathcal{D}$ is the sample space.

## III. Experiments and Results

Several experiments on publicly available datasets and comparisons have been carried out, using baseline and state-of-the-art classifiers. The datasets, experimental setup, and the results obtained along with explainability validation tests are provided in the rest of this section.

### A. Dataset Description

Experiments were conducted on both general-purpose UCI benchmark datasets, *i.e.*, Seismic Bumps, Sonar, and Ionosphere [14], as well as on a simulated maritime dataset modeling a Combined Diesel Electric and Gas (CODLAG) propulsion system installed on a naval frigate [15]. The CODLAG dataset models two key coefficients: the gas turbine compressor (GTC) degradation coefficient ($k_c$) and the gas turbine (GT) degradation coefficient ($k_t$). Based on [3], and under the assumption of a 2000h per service time per year, we apply thresholds to $k_c$, $k_t$ to transform the problem into a binary classification setting, defining a compressor (6) and a turbine (7) health state:

$$y_{k_c} = \begin{cases} 1, & 0.95 \le k_c < 0.98 \\ 0, & 0.98 \le k_c \le 1.00 \end{cases} \quad (6)$$

$$y_{k_t} = \begin{cases} 1, & 0.97 \le k_t < 0.99 \\ 0, & 0.99 \le k_t \le 1.00 \end{cases} \quad (7)$$

### B. Experimental Setup

The performance of the proposed framework was compared to Support Vector Machine (SVM), Logistic Regression (LR), Classification and Regression Tree (CART) algorithm [16], CatBoost algorithm [17], as well as the state-of-the-art T2g-Former [18] transformer-based model. A 10-fold cross-validation strategy was adopted for a fair comparison performance of all classifiers. The ranges of the hyperparameters tested in the case of the SVM classifier, were $C \in [10^{-3}, 10^{3}]$, $\gamma \in [10^{-5}, 10^{5}]$ and for LR, $C \in [10^{-4}, 10^{2}]$. For tree-based models, the maximum tree depth for CART was tuned within the range [5, 20], for CatBoost, the tree depth was optimized within [4, 14]. Neural network-based models were tuned with respect to the number of layers, hidden layer sizes, and embedding dimensions, with values selected from the intervals [1, 4], [64, 256], and [128, 256], respectively.

### C. Results

Table I summarizes the comparisons performed in terms of the average AUC-ROC (Area Under the Receiver Operating Characteristic Curve) value. The proposed framework achieves competitive accuracy on most of the datasets, while maintaining state-of-the-art status in the naval propulsion dataset. Specifically, the proposed framework achieves 99% in terms of the AUC value for both the GT and GTC coefficients, with a mean confidence level ranging within [0.07, 0.39]. The proposed framework provides a good tradeoff between performance and explainability in the sense that it has comparable results with current black-box classifiers. However, it has to be mentioned that an advantage over the rest compared approaches is that the proposed framework is explainable. Therefore, in addition to predictive performance, the quality of the generated explanations was quantitatively assessed through validation experiments, using synthetic data presented in the next section.

TABLE I. RESULTS IN TERMS OF AUC

| Model/ Dataset | Seismic Bumps | Sonar | Ionosphere | Naval dataset (GTC / GT) | |
|---|---|---|---|---|---|
| LR | 0.71 | 0.80 | 0.84 | 0.93 | 0.95 |
| SVM | 0.69 | 0.87 | 0.94 | 0.99 | 0.99 |
| CART [16] | 0.73 | 0.79 | 0.88 | 0.95 | 0.91 |
| CATboost [17] | 0.71 | 0.85 | 0.92 | 0.94 | 0.93 |
| T2g-Former [18] | 0.66 | 0.67 | 0.82 | 0.97 | 0.93 |
| **Proposed framework** | 0.75 | 0.85 | 0.94 | 0.99 | 0.99 |

TABLE II. COMPARISONS OF NNS WITH AND WITHOUT THE FUZZY CLASSIFIER HEAD IN TERMS OF EXPLANATION VALIDATION METRICS.

| Metric | RC NN | FA NN | JI NN | RC NN-FCH | FA NN-FCH | JI NN-FCH |
|---|---|---|---|---|---|---|
| G | 0.71 | 0.88 | 0.83 | **0.71** | **0.96** | **0.95** |
| GxI | 0.55 | 0.64 | 0.52 | 0.64 | 0.74 | 0.64 |
| IG | 0.60 | 0.68 | 0.56 | 0.68 | 0.80 | 0.72 |
| SG | 0.71 | 0.85 | 0.79 | **0.71** | **0.95** | **0.94** |
| SIG | 0.59 | 0.67 | 0.55 | 0.69 | 0.80 | 0.72 |
| EG | 0.54 | 0.60 | 0.47 | 0.68 | 0.79 | 0.71 |

### D. *Explainability Validation Tests*

Relying solely on visual assessment of explanation results can be misleading. Therefore, we follow the sanity checks proposed in [11] to test our framework, and construct a set of synthetic logistic regression (LR) datasets in which the important features are explicitly controlled. By explicitly defining the ground-truth of feature importance, these synthetic LR datasets provide a reliable benchmark for assessing the faithfulness of the generated explanations. Regarding the evaluation metrics, the Jaccard index (JI) was utilized for raw feature overlap, Rank Correlation (RC) to measure agreement in feature ordering and Feature Agreement (FA) to identify consistency between the features. For each experiment, 10 randomly generated LR datasets were constructed. More specifically, 10,000 i.i.d. feature vectors $X \in \mathbb{R}^d$ were sampled from a normal distribution, and a sparse ground-truth weight vector $w \in \mathbb{R}^d$ was constructed by selecting three distinct informative features and assigning them nonzero weights. The saliency maps were evaluated with several gradient-based attribution methods, such as plain gradients (G), gradient ⊙ input (GxI), integrated gradients (IG), SmoothGrad (SG), smoothGrad-IG (SIG) and expected gradients (EG). To clarify the importance of the Fuzzy Classifier Head (FCH), comparisons were performed, among a NN combined with a FCH, indicated as NN-FCH, and the baseline NN model, *i.e.*, without the FCH. As presented in Table II, the evaluation is performed on 6 types of gradient attribution methods for tabular data.

The two best triplet combinations of the performing gradient attribution methods and models are highlighted in bold. The results indicate that plain gradients (G) and SG achieve the best performance across both models. This first indicates that simple gradient-based saliency methods are sufficient to reliably recover relevant features in the examined setting and are well aligned with the underlying model behavior. Second, incorporating fuzzy logic into the prediction level improves the quality of the explanations by approximately 8%, with average JI and FA exceeding 94%. Regarding the FDT explanations, the average JI and FA values are above 90%. These results confirm an agreement between the gradient-based saliency explanations, the features identified by the FDT, and the ground truth, supporting the reliability of the proposed framework. Furthermore, model weight randomization and cascading randomization sanity checks were applied [11], which validated that the inferred explanations are not random, being sensitive to the learned model.

## IV. EXPLAINABLE EXAMPLES

To better understand the explainability of the proposed neuro-fuzzy prediction framework, in this section, two examples related to early fault detection in naval propulsion systems [15], are provided. The examples correspond to a healthy and degraded compressor state under a setting of lever position configuration $l = 9.3$ (representing an intense engine state). Considering a healthy state of an NPS, it corresponds to normal operating conditions, where GTC and gas turbine (GT) have not degraded; this corresponds to the predefined $y_{k_c}$and $y_{k_t}$ health states (6), (7) being equal to zero. The two samples were randomly selected from the dataset. Each sample is characterized by the following six features: i) rate of revolutions of the gas generator (($GGn$) measured in Rpm), ii) gas turbine exhaust gas pressure (($Pexh$) measured in bar), iii) outlet air temperature of the compressor (($T2$), measured in ºC), iv) air pressure (($P2$) measured in bar), v) fuel flow (($mf$) measured in kg/s), vi) port propeller torque (($Tp$) measured in kN). The first sample is related to a Healthy Compressor State (HCS) and is characterized by $GGn = 9720.441 rpm$ , $Pexh = 1.051 bar$ , $P2 = 22.756 bar$ , $mf = 1.828 kg/s$, and $Tp = 644.81 kN$. The second sample represents a Degraded Compressor State (DCS) characterized by $GGn = 9780.103 rpm,$ $Pexh = 1.049 bar$ , $T2 = 788.806 C$, $P2 = 22.889 bar$, $mf = 1.762 kg/s$ and $Tp = 645.076 kN$**.** The value ranges for the linguistics that are used to characterize the parameters are the following: a) for $T2$ parameter, Low: [766, 786], Medium: [773, 793], High: [778,798]; b) $Pexh$ parameter, Low: [1.0475, 1.0505], Medium: [1.0485, 1.0515], High: [1.0495, 1.0525]; c) *GGn* parameter, Low: [9700, 9760], Medium: [9720, 9780], High: [9740, 9800]. Using the explainable component, for the first sample, the three most important features are extracted in the form of saliency maps, which are the following: *Pexh* (0.60), *P2* (0.51), *Tp* (0.10). Saliency scores indicate the relative importance of each feature in the prediction process within each sample, with Pexh (0.60) and P2 (0.51) identified as the most influential features. Additionally, the local explanations derived from the FDT, in the form of rules are the following:

Rule 1 (HCS): **IF** GGn is Low **AND** Pexh is (Medium **OR** High) **THEN** the Compressor State is Healthy ( $s = 42.88, p = 0.98$).

Rule 2 (HCS): **IF** GGn is (Low **OR** Medium) **AND** Pexh is High **THEN** the Compressor State is Healthy ( $s = 38.37, p = 0.92$).

Similarly, for the second example, the three most important features from the saliency maps are *P2* (0.72), *Pexh* (0.55), *mf* (0.26). The resulting rules are the following:

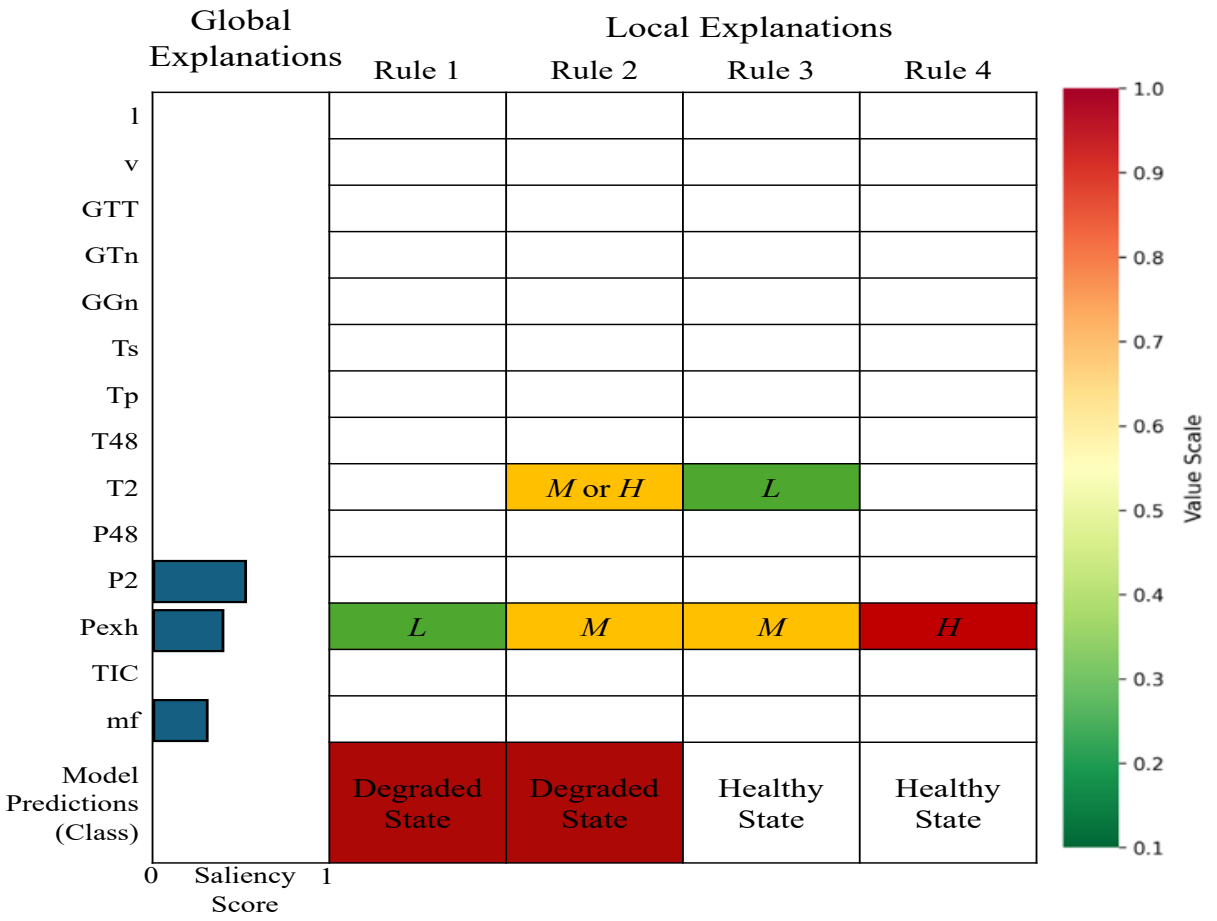

Fig. 3. The aggregated explanations for the DCS example

Rule1(DCS): **IF** *Pexh* is *Low* **THEN** the Compressor State is *Degraded* ($s = 68.07, p = 1$).

Rule 2(DCS): **IF** *Pexh* is *Medium* **AND** *T2* is (*Medium* ***OR*** *High)* **THEN** the Compressor State is *Degraded* ( $s = 9.8, p = 0.85$).

A visualization of the aggregated output of the explainable component, which combines the predictions of the model with gradient-based saliency feature importance maps and the extracted rules is presented in Fig. 3. In this figure, the feature-attribution explanations are on the left, where the three most important features are illustrated in bar chart. The bar lengths are proportional to the calculated saliency scores (3). In this example, the most important features are $P2$, $Pexh$ and $mf$. Furthermore, the local explanations derived from the FDT are presented, using linguistic terms, *i.e.*, (L)ow, (M)edium, (H)igh that corresponds to the calculated values. The most influential fuzzy rules associated locally with degraded operating states prominently involve exhaust temperature ($T2$) and pressure-related features, *i.e.*, $Pexh$. It should be noted that the FDT eliminates the remaining irrelevant features, which in this example are 12 in total and are depicted in white box (Fig. 3). The explanations derived from the proposed framework are consistent with the existing literature [15], confirming their validity, which reports that compressor and turbine fouling lead to increased fuel consumption and higher exhaust gas temperatures, due to reduced compressor efficiency and airflow degradation.

## V. Conclusions

This paper presented a novel neuro-fuzzy prediction framework comprising a fuzzy classifier based on neural networks, which serves as the primary predictive model with a two-stage explainable component. To the best of our knowledge, this is the first framework based on fuzzy logic that enables both feature-attribution and local rule-based explanations of black-box models. The main contributions emerging include the following: a) it employs an explainable component that explicitly models uncertainty, while maintaining end-to-end trainability and competitive performance; b) it provides a good tradeoff between performance and explainability in the sense that it has comparable results with current black-box classifiers, while offering the advantage of explainability; c) it is capable of automatically extracting rules directly from data, without requiring prior domain knowledge, thus limiting the reliance on manual rule engineering. Evaluations on synthetic datasets validated the saliency maps and had 8% quality improvement of explanations, when fuzzy logic was incorporated at the prediction level. The experimental results indicated that the proposed framework can provide predictions outperforming both conventional and state-of-the-art approaches. Future work will focus on extending the proposed framework to incorporate temporal dependencies and its application on other scientific domains, including health and aviation.

## References

[1] D. D. M. Frangopol, D. P. Bocchini, A. Decò, D. S. Kim, D. K. Kwon, D. N. M. Okasha, and D. Saydam, “Integrated life-cycle framework for maintenance, monitoring, and reliability of naval ship structures,” *Naval Engineers Journal*, vol. 124, no. 1, pp. 89–99, 2012.

[2] K. Kwon, D. M. Frangopol, and S. Kim, “Fatigue performance assessment and service life prediction of high-speed ship structures based on probabilistic lifetime sea loads,” *Structure and Infrastructure Engineering*, vol. 9, no. 2, pp. 102–115, 2013.

[3] F. Cipollini, L. Oneto, A. Coraddu, A. J. Murphy, and D. Anguita, “Condition-based maintenance of naval propulsion systems: Data analysis with minimal feedback,” *Reliability Engineering & System Safety*, vol. 177, pp. 12–23, 2018.

[4] J. Gao, S. Dong, J. Cui, M. Yuan, and J. Zhao, “DRN-GAN: an integrated deep learning-based health degradation assessment model for naval propulsion system,” *Engineering Computations*, vol. 39, no. 6, pp. 2306–2325, 2022.

[5] F. Javadnejad, H. J. Park, S. Kovacic, and A. Sousa-Poza, “A Novel BiGMM-HMM Framework for Predictive Maintenance in Naval Vessel Propulsion Equipment,” in *2024 19th Annual System of Systems Engineering Conference (SoSE)*, 2024, pp. 116–123.

[6] M. A. Sahraoui, C. Rahmoune, A. Damou, F. Gougam, and A. Afia, “Advancing condition-based maintenance of naval propulsion systems with ensemble learning techniques,” *Advances in Mechanical Engineering*, vol. 16, no. 11, p. 16878132241298373, 2024.

[7] J.-S. Jang, “ANFIS: adaptive-network-based fuzzy inference system,” *IEEE transactions on systems, man, and cybernetics*, vol. 23, no. 3, pp. 665–685, 1993.

[8] R. Das, S. Sen, and U. Maulik, “A survey on fuzzy deep neural networks,” *ACM Computing Surveys* vol. 53, no. 3, pp. 1–25, 2020.

[9] M. Yeganejou, R. Kluzinski, S. Dick, and J. Miller, “An end-to-end trainable deep convolutional neuro-fuzzy classifier,” in *2022 IEEE Conference on Fuzzy Systems (FUZZ-IEEE)*, 2022, pp. 1–7.

[10] M. Yeganejou, K. Honari, R. Kluzinski, S. Dick, M. Lipsett, and J. Miller, “Dcnfis: Deep convolutional neuro-fuzzy inference system,” *arXiv preprint arXiv:2308.06378*, 2023.

[11] J. Adebayo, J. Gilmer, M. Muelly, I. Goodfellow, M. Hardt, and B. Kim, “Sanity checks for saliency maps,” *Advances in neural information processing systems*, vol. 31, 2018.

[12] C. Olaru and L. Wehenkel, “A complete fuzzy decision tree technique,” *Fuzzy sets and systems*, vol. 138, no. 2, pp. 221–254, 2003.

[13] C. Z. Janikow, “Fuzzy decision trees: issues and methods,” *IEEE Transactions on Systems, Man, and Cybernetics, Part B (Cybernetics)*, vol. 28, no. 1, pp. 1–14, 1998.

[14] C. L. Blake, “UCI repository of machine learning databases,” *http://www. ics. uci. edu/ mlearn/MLRepository. html*, 1998.

[15] A. Coraddu, L. Oneto, A. Ghio, S. Savio, D. Anguita, and M. Figari, “Machine learning approaches for improving condition-based maintenance of naval propulsion plants,” *Proceedings of the Institution of Mechanical Engineers, Part M: Journal of Engineering for the Maritime Environment*, vol. 230, no. 1, pp. 136–153, 2016.

[16] T. Hastie, R. Tibshirani, and J. Friedman, “An introduction to statistical learning,” 2009.

[17] L. Prokhorenkova, G. Gusev, A. Vorobev, A. V. Dorogush, and A. Gulin, “CatBoost: unbiased boosting with categorical features,” *Advances in neural information processing systems*, vol. 31, 2018.

[18] J. Yan, J. Chen, Y. Wu, D. Z. Chen, and J. Wu, “T2g-former: organizing tabular features into relation graphs promotes heterogeneous feature interaction,” in *Proceedings of the AAAI Conference on* , 2023, vol. 37, no. 9, pp. 10720–10728.